\documentclass[10pt,twocolumn,letterpaper]{article}

\usepackage{wacv}
\usepackage{times}
\usepackage{epsfig}
\usepackage{graphicx}
\usepackage{amsmath}
\usepackage{amssymb}
\usepackage{balance}

\usepackage[table]{xcolor}
\usepackage{caption}

\wacvfinalcopy 

\def\wacvPaperID{193} 

\ifwacvfinal\fi

\begin{document}

\title{An end-to-end generative framework for video segmentation and recognition}

\author{Hilde Kuehne \\
              University of Bonn \\
{kuehne@iai.uni-bonn.de}
\and
Juergen Gall \\
              University of Bonn \\
{gall@iai.uni-bonn.de}
\and
Thomas Serre \\
              Brown University \\
{thomas\_serre@brown.edu}
}

\maketitle

\begin{abstract}

We describe an end-to-end generative approach for the segmentation and recognition of human activities. In this approach, a visual representation based on reduced Fisher Vectors is combined with a structured temporal model for recognition. We show that the statistical properties of Fisher Vectors make them an especially suitable front-end for generative models such as Gaussian mixtures. The system is evaluated for both the recognition of complex activities as well as their parsing into action units. Using a variety of video datasets ranging from human cooking activities to animal behaviors, our experiments demonstrate that the resulting architecture outperforms state-of-the-art approaches for larger datasets, \ie when sufficient amount of data is available for training structured generative models.

\end{abstract}

\section{Introduction}

\enlargethispage{.3cm}

The growing need for automated video monitoring and surveillance systems is quickly reshaping our research landscape. Much of the current research on action recognition has focused on semi-realistic problems such as categorizing short clips consisting of one single action (\eg kick, pour, throw, pick). However, many real-world applications will require methods that can solve more realistic problems including the recognition and parsing of complex activities in long continuous recordings, often consisting of sequences of goals and sub-goals. 

Most successful approaches to action recognition have typically relied on unstructured models of video sequences. A holistic visual representation is usually computed over an entire video clip and then passed to a discriminative classifier to yield a single categorization label per video. These methods have been successful for the recognition of single-action video clips (see \eg~\cite{Wang2013Action}). However, they do not appear to be well suited for the recognition of daily activities that require the modeling of complex behavior sequences. 

Several extensions of these unstructured models have been proposed to try to address this challenge. One popular approach relies on sliding (temporal) windows whereby videos are decomposed into a sequence of shorter segments that can be individually classified with discriminative approaches~\cite{Rohrbach2012database, Cheng2014Temporal,Bhattacharya2014Recognition}. However, these approaches have, for the most part, only been tested on a handful of relatively small datasets that do not capture the rich and diverse nature of daily activities. As we will show, these approaches are not competitive on more challenging activity datasets.


\begin{figure}
\centering
\includegraphics[width=0.95\linewidth]{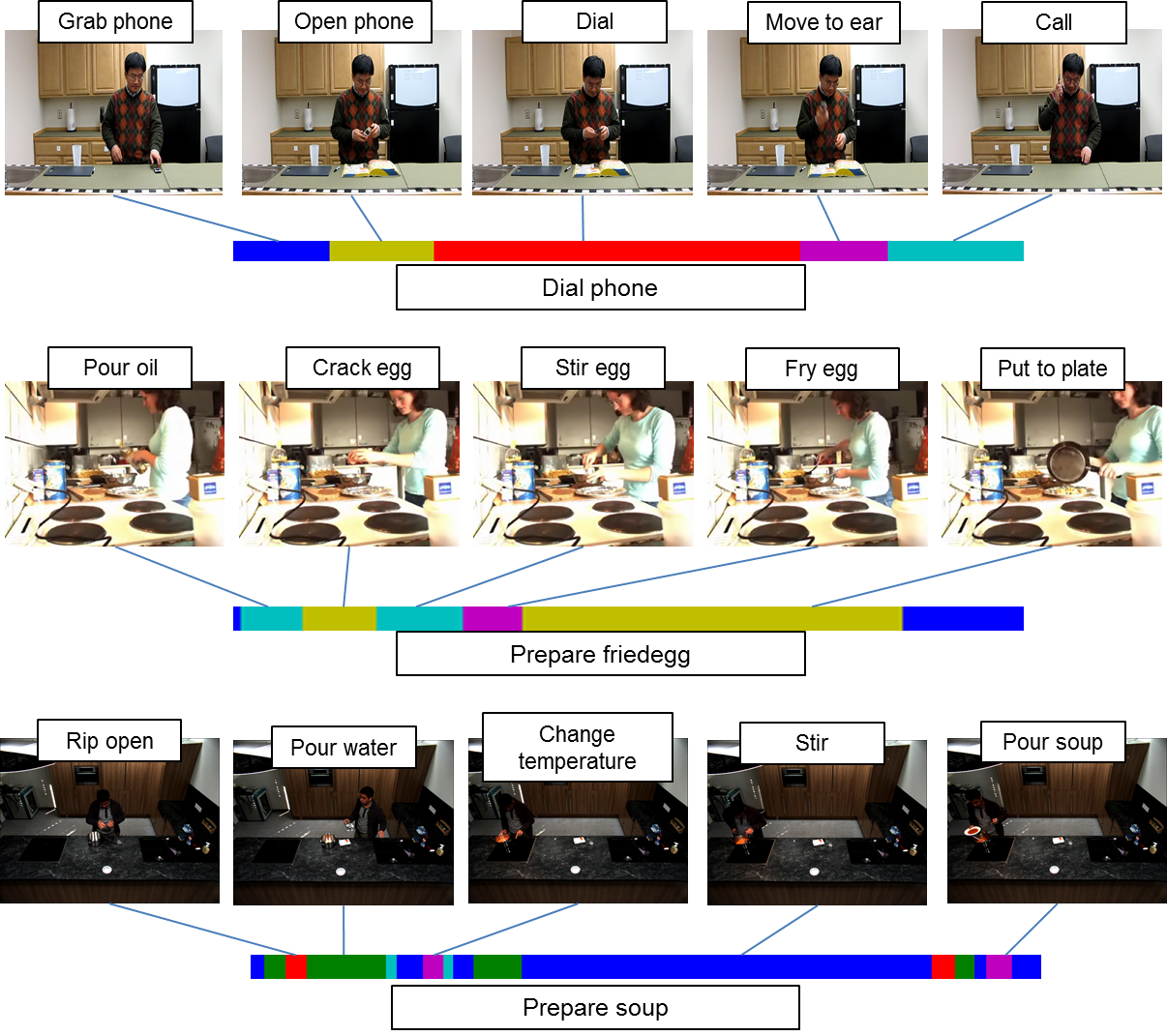}   \\
\caption{Segmentation and recognition of human activities with a) the ADL dataset (``dial phone''), b) the Breakfast dataset (``prepare fried eggs'') and c) the MPII cooking dataset (``prepare soup'').}
\label{fig:exp_action_units}\vspace{-3mm}
\end{figure}

Structured temporal models, on the other hand, have reached an impressive level of maturity in several engineering domains and speech recognition~\cite{Young2006Htkbook} in particular. These models would appear more appropriate than their unstructured counterparts for the recognition of human activities. Somewhat surprisingly, relatively little effort has been devoted to adapting these approaches to human action recognition  (but see \eg~\cite{ChiaChih2011Modeling,Kuehne2014Language}). One of the main reasons why structured generative methods have not found more widespread acceptance in action recognition is that, unlike for speech analysis where large annotated corpora are available, video databases have been comparatively limited in size~\cite{Kuehne2014Language}. 

With the emergence of larger video datasets (\eg CRIM13~\cite{Burgos12CRIM} and Breakfast~\cite{Kuehne2014Language}), these models are more likely to start exhibiting competitive performance. For instance, encouraging results were obtained in~\cite{Kuehne2014Language} using Hidden Markov Models (HMMs) combined with a context-free grammar to learn cooking activities. One of the main limitations associated with standard HMM toolboxes (such as the HTK used in~\cite{Kuehne2014Language}) is the use of Gaussian mixtures, which typically require input data to be normally-distributed as well as low-dimensional to prevent overfitting. Standard visual representations such as Bag-of-Words or Fisher Vectors (FVs) thus constitute a poor choice for HMMs and other generative approaches, because they typically yield sparse and high-dimensional visual representations.

Here, we describe an approach for the construction of reduced FVs which is particularly amenable to structured temporal models. FVs have been shown to achieve state-of-the-art accuracy in action recognition~\cite{Peng2014Stacked}. They have also been shown to maintain good classification accuracy when used in conjunction with dimension reduction techniques~\cite{Csurka2011Fisher,Jegou2012Aggregating}. Hence, this makes them good candidates for modeling by Gaussian mixtures. As we will show, the proposed approach yields a very substantial improvement in recognition accuracy on a variety of activity segmentation and recognition tasks, ranging from the recognition of human daily activities to the segmentation of rodent social interactions.

To summarize, we describe an approach to improve the efficiency of state-of-the-art feature encoding methods~\cite{Csurka2011Fisher,Jegou2012Aggregating} that are especially amenable to generative models. We systematically evaluate the proposed approach using a variety of standard activity datasets and demonstrate significant improvements for datasets that contain sufficient training data. 



\section{Related work}

\subsection{Fisher vectors} 

Fisher kernel methods were originally proposed as a way to derive kernels for discriminative classifiers from generative models~\cite{Jaakkola1999Exploiting}. They were later adapted to represent feature sets used for image classification~\cite{Perronnin2007Fisher}. The application of an $L_2$ norm and power normalizations combined with a method for sampling FVs based on a spatial pyramid were then shown to significantly improve their accuracy~\cite{Perronnin2010Improving}. More recently, FVs have been shown to yield not only higher classification accuracy, but also much more compact feature vectors~\cite{Jegou2012Aggregating}. 

The application of FVs to action recognition was first explored in~\cite{Wang2012Comparative}, where the authors used a standard video descriptor (HOGHOF) to compare different encoding methods on two different datasets. It was shown that FVs often outperform other methods, a result that was further replicated in a separate study~\cite{Sun2013Large}. The combination of FVs and Dense Trajectory Features (DTFs) was also demonstrated to work exceedingly well for the recognition of actions~\cite{Wang2013Action,Oneata2013Action}. All the aforementioned approaches are based on  discriminative classification methods trained on (short) single-action pre-segmented video clips. We are not aware of previous work focusing on the statistical properties of FVs in the context of a generative action recognition models.

\subsection{Structured temporal models} 

Most early approaches for action recognition with structured temporal models relied on either motion capture data~\cite{Guerra2005Discovering,Sminchisescu2005Conditional} or hand-labeled trajectories~\cite{Rao2002ViewInvariant}. Several temporally structured models have been applied since on video data including generative mixture models~\cite{Messing2009Activity}, Bayes Networks~\cite{Ryoo2009Human} and an HMM/SVM combination~\cite{ChiaChih2011Modeling}. 


More recent work has focused on the problem of detecting and segmenting human activities in videos. In~\cite{Si11Unsupervised}, a semantic scene label map was built as context for agent actions to automatically learn AND-OR grammars from videos. In~\cite{Bhattacharya2014Recognition}, Linear Dynamical Systems theory was used to detect events in complex video datasets. Long-term relations were also considered in the ``sequence memorizer'' described in~\cite{Cheng2014Temporal}, which uses a Bayesian nonparametric model to simultaneously detect and classify events within a video stream. A similar idea is proposed in the work of~\cite{Kuehne2014Language}, using a context free grammar in combination with HMMs to model longer temporal sequences of smaller action units.    
In~\cite{Fathi13Modeling}, activity models were based on the detection of changes in state-specific regions of interest (\eg the lid of a coffee jar for 'opening coffee jar' and 'closing coffee jar' actions). The authors used SVM-based state detectors to detect the beginning and end of short task-oriented action units such as ``hold spoon'' or ``stir coffee''. 

A higher-level representation based on stochastic context-free grammars was used in~\cite{Vo2014Stochastic} where body pose information (\ie hand positions) was used for classification. A closely related approach was proposed in~\cite{Pirsiavash2014Parsing} where action units were combined with a set of production rules to build a grammar to model the hierarchical temporal structure of human activities. The system was able to learn and parse action units derived from the Olympic sport dataset. 

Here, we build on our earlier work ~\cite{Kuehne2014Language} using HMMs combined with a simple grammar to model complex human activities as sequences of action units.

\section{System description}

\subsection{Fisher vectors}
\label{sec:Fisher_vector_encoding}

We briefly review the key steps involved in FV computation and frame-based action recognition. We refer the reader to~\cite{Sanchez2013Image} for a more detailed description. The main assumption behind FVs is that local feature descriptors may be modeled by a probability density function. Here, we consider a Gaussian mixture Model (GMM) with $K$ components defined by the associated mixture weights, mean vectors ${\mu_k}$ and variances ${\sigma_k}$. FVs characterize how a feature set $\mathrm{X} =\{ x_t | t = 1, \dots, T\}$ deviates from a learned distribution. For each feature set $\mathrm{X}$, the resulting gradients $\mathcal{G}_{\mu_k}^\mathrm{x}$ and $\mathcal{G}_{\sigma_k}^\mathrm{x}$ each have the dimensionality $D$ of the original feature descriptor and they are computed for each mixture of the GMM as described in~\cite{Csurka2011Fisher}.

The concatenation leads to an overall $2\times D \times K$ dimensional FV representation $\hat{x}$ of the original feature set $X$ with $\hat{x} = [\mathcal{G}_{\mu, k}^\mathrm{x} , \mathcal{G}_{\sigma, k}^\mathrm{x} ]'$. Following~\cite{Perronnin2010Improving}, we applied an L2-normalization to these vectors. Additionally, the authors in ~\cite{Perronnin2010Improving} observed that the more Gaussian components are used, the sparser the FVs become. We followed their suggestion to use a power normalization (\mbox{$g(\hat{x}) = sign(\hat{x}) \sqrt{\hat{x}}$})  to reduce the sparsity of the FVs. As the resulting FVs are too high dimensional to be processed in a generative framework, we used PCA to reduce the overall dimensionality of the feature vector~\cite{Jegou2012Aggregating} and to further whiten the data. 

\begin{figure}[t!]
\centering
\includegraphics[width=0.6\linewidth]{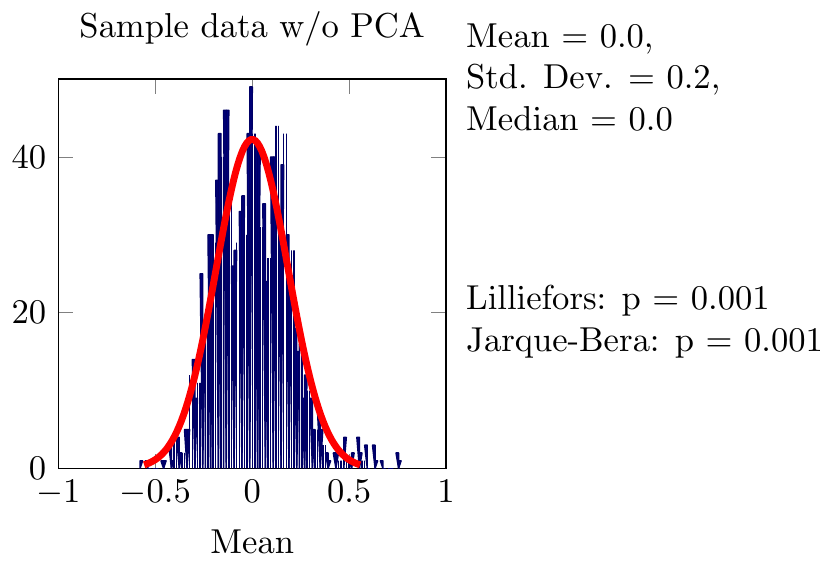} 
\includegraphics[width=0.6\linewidth]{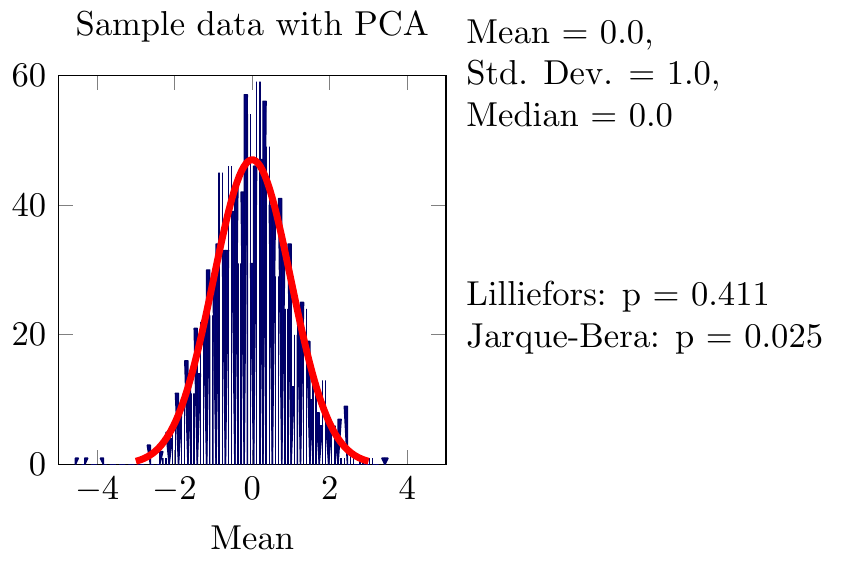}
\includegraphics[width=0.6\linewidth]{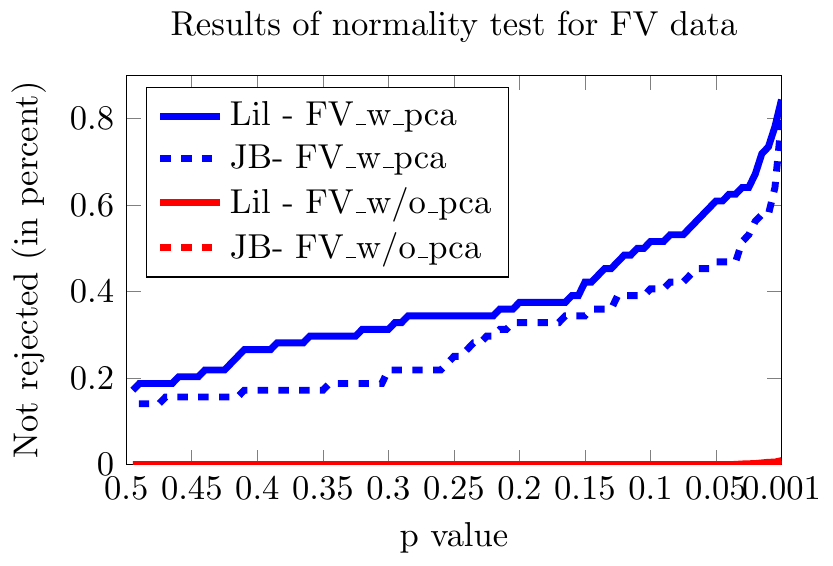}
\caption{Distribution of FV samples before and after PCA and results of normality test (Lil = Lilliefors, Jb = Jarque-Bera) with decreasing significance levels for FV samples before and after PCA.}\label{fig:exp_normalityTest}
\end{figure}

\subsection{Normality test}
\label{sec:Normality_test}

The HTK recognition framework used here (see section~\ref{sec:Generative_recognition}), like most other systems for automated speech recognition, relies on HMMs with observation probabilities modeled by Gaussian mixtures. Higher dimensional Gaussian mixtures are prone to overfitting, especially when given only a limited amount of training data. This can be compensated to a certain extent by reducing the number of mixtures used. In general, we found that best results were obtained with one Gaussian per state which is consistent with the practice reported in ~\cite{Kuehne2014Language}. It is thus highly desirable for input data to be  normally distributed.

In order to test the normality of FVs for video data, we considered different normality tests. To evaluate how dimensionality reduction using PCA affects the normality of the resulting feature vector, we randomly sampled data along each dimension of the feature vectors and test the skewness and kurtosis of the resulting distributions using the Lilliefors~\cite{Lilliefors1967Kolmogorov} and the Jarque-Bera test~\cite{Jarque1987atest}, respectively. 
We tested the null hypothesis that a given dimension is normally distributed and estimated the number of dimensions for which the null hypothesis is valid (for decreasing significance levels in the range 0.5--0.001). We applied this test to FV samples before and after PCA. 
Results shown in Figure~\ref{fig:exp_normalityTest} confirm that PCA yields distributions that are closer to a normal distribution. For instance, at a significance level of $\alpha=0.001$, a mere 0.53\% of the original FV dimensions pass the Lillifors test (none for Jaque-Bera), whereas 84.3\% (79.6\% Jaque-Bera) of the PCA-reduced data dimensions pass significance. This is quite evident already when we consider the first dimension of the feature vector (before and after PCA), as shown in Figure~\ref{fig:exp_normalityTest}. For comparison, we also tested the BoWs as used in our previous work~\cite{Kuehne2014Language}. Here, the null hypothesis was always rejected, irrespective of the significance level, suggesting that none of the dimensions are normally distributed.

Overall, PCA helps to build a feature vector that better fits the normality assumption of the proposed HMM-based model. As we will show in Section~\ref{sec:System_evaluation}, this yields significant gains in activity recognition accuracy.

\begin{figure*}[t!]
\begin{center}
\includegraphics[width=.99\linewidth]{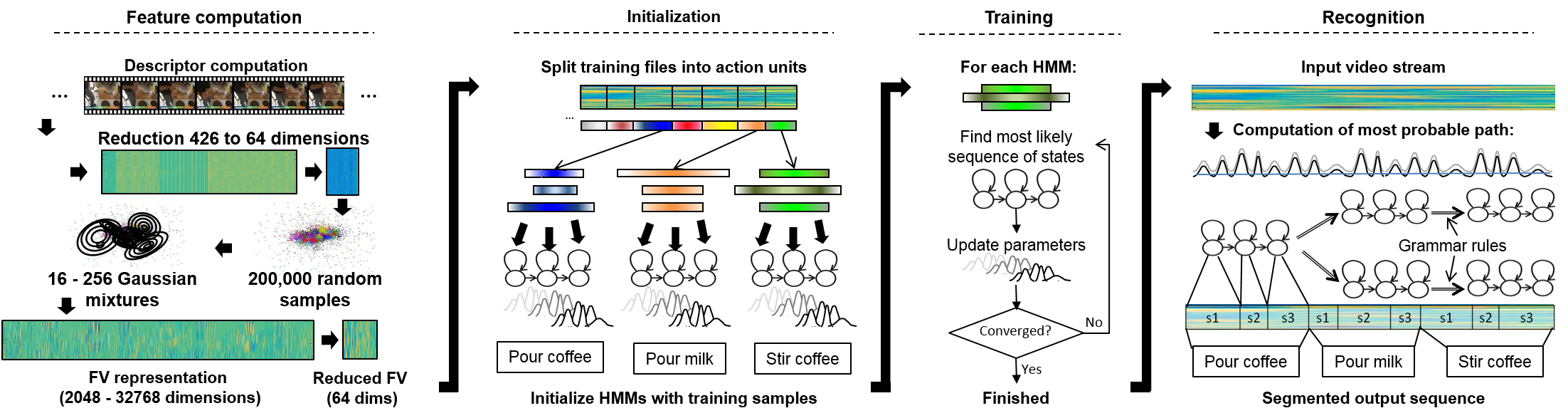}
\end{center}
\caption{Overview of the recognition pipeline: DT features are computed and the corresponding descriptor is reduced to 64 dimensions. A total of 200,000 features are randomly sampled and fitted to GMMs ($K= 16, 32, 64, 128$ or 256). An FV representation is computed for each frame of the video. The corresponding representation is further reduced from 2048 -- 32,768 down to 64 dimensions. During training, each HMM is initialized with action unit samples. State boundaries are re-estimated and the GMMs are updated according to the new state boundaries until convergence. During recognition, HMMs are combined with a learned context-free grammar and the most probable sequence of action units is determined. }\label{fig:system_overview}
\end{figure*}

\subsection{A generative recognition pipeline}
\label{sec:Generative_recognition}

In the following, we briefly give an overview of the pipeline used (see Figure~\ref{fig:system_overview}). We used an improved version~\cite{Wang2013Action} of the Dense Trajectory Features (DTFs)~\cite{Wang2013Dense} for datasets with camera motion. The dimensionality of the feature descriptors was first reduced from 426 dimensions to 64 dimensions by PCA, following the procedure described in~\cite{Oneata2013Action}. 


We sampled 200,000 random features to fit the GMMs. FVs were computed using 50,000 frames sampled from the training data. For each reference frame, FVs were computed over a 20-frames sliding window. The dimensionality of the resulting vector was further reduced to 64 dimensions using PCA (see section~\ref{sec:Fisher_vector_encoding}). Thus, each frame is then represented by a 64-dimensional FV. We further applied an L2-normalization to each feature dimension separately for each video clip.

The proposed recognition system contains two main components: a set of HMMs is used to model all possible action units found in the dataset and a grammar is used to model possible sequences of those units. The number of hidden states for each HMM was set to $1/10$ of the mean length  of the corresponding action units. All HMMs are based on a left-to-right feed-forward topology, allowing only self-transitions and transitions to the next state. The initial state transitions probabilities were set to default values (self: $p=0.6$, next $p=0.4$). To initialize the state distribution, we subdivided each action unit evenly over time and associated each subdivision to a hidden state. Thus, frames at the beginning or end of an action unit get always associated to the first and last states due to the left-to-right topology.


During training,  unit states were re-estimated using the Baum-Welch algorithm, \ie by finding the HMM parameters that maximise the probability of a given set of observations. For details concerning the training and recognition with HTK, we refer the reader to to~\cite{Young2006Htkbook,Kuehne2014Language} for details. As the number of samples per class or in our case, per action unit follows a long-tail distribution, with few classes being frequent and a large number of classes being relatively rare, we enforced a minimum and maximum number of training samples (see Table~\ref{tab:overview_datasets}) for a balanced training data set across classes. When needed, artificial samples were generated by synthetic minority over-sampling to guarantee a minimum number of samples.

During recognition, we followed the approach described in by~\cite{Kuehne2014Language} formalizing activity recognition and segmentation as the problem of finding the most probable sequence of action units from an observed input sequence. A context free grammar was built automatically using available annotations.  For the CRIM13 dataset~\cite{Burgos12CRIM}), we favored a bi-gram model which defines the transition probability to the next possible units instead of absolute paths. This is a richer model and it is more appropriate for modeling animal behavior which tends to be relatively stochastic compared to the human activities found in other datasets. 

The Viterbi algorithm was used to find the most probable sequence of action units. The output of the algorithm includes the best matching sequence of action units, their beginning and end frames, and the corresponding observation probabilities (see~\cite{Young2006Htkbook}).

\section{Evaluation}

\subsection{Datasets}
\label{sec:Datasets}

\begin{figure*}[t!]
\centering
a)\includegraphics[width=0.22\textwidth]{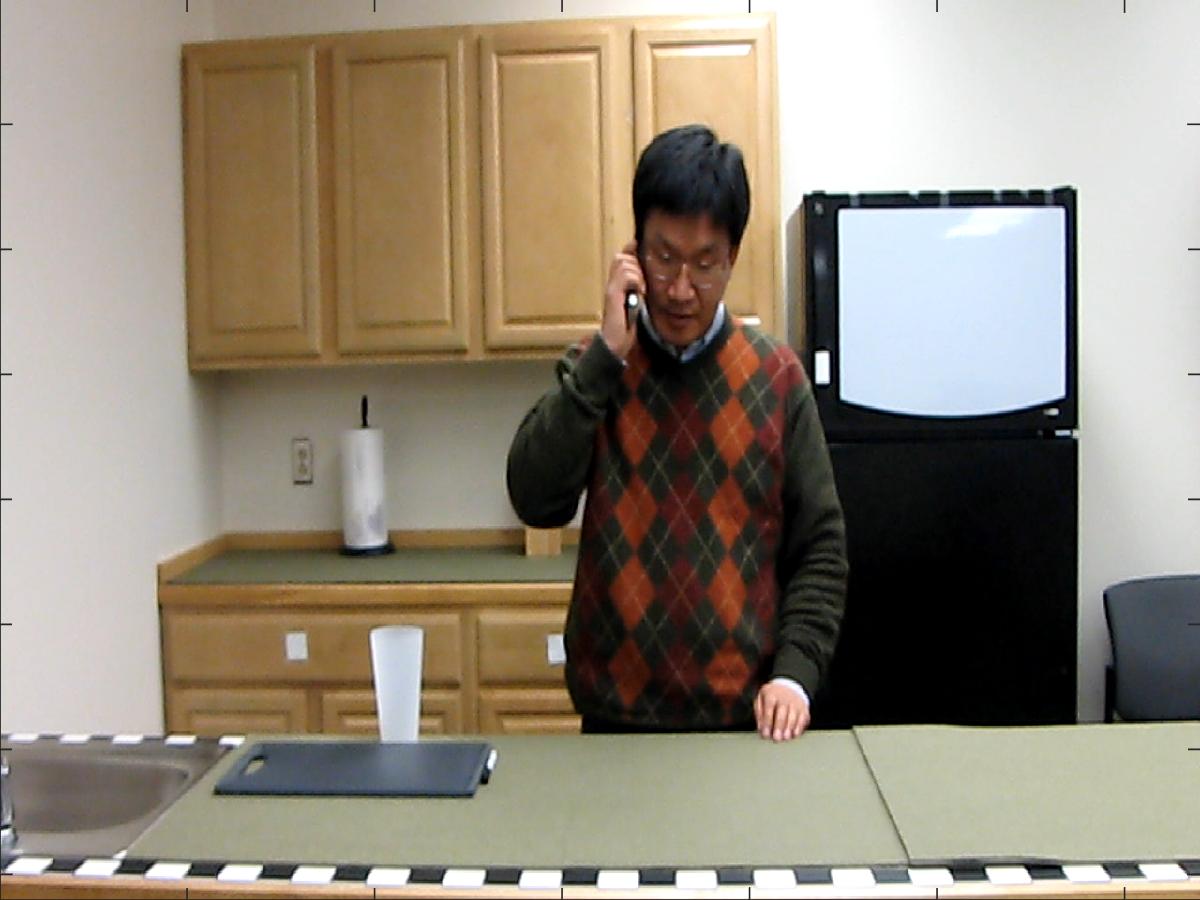}
b)\includegraphics[width=0.22\textwidth]{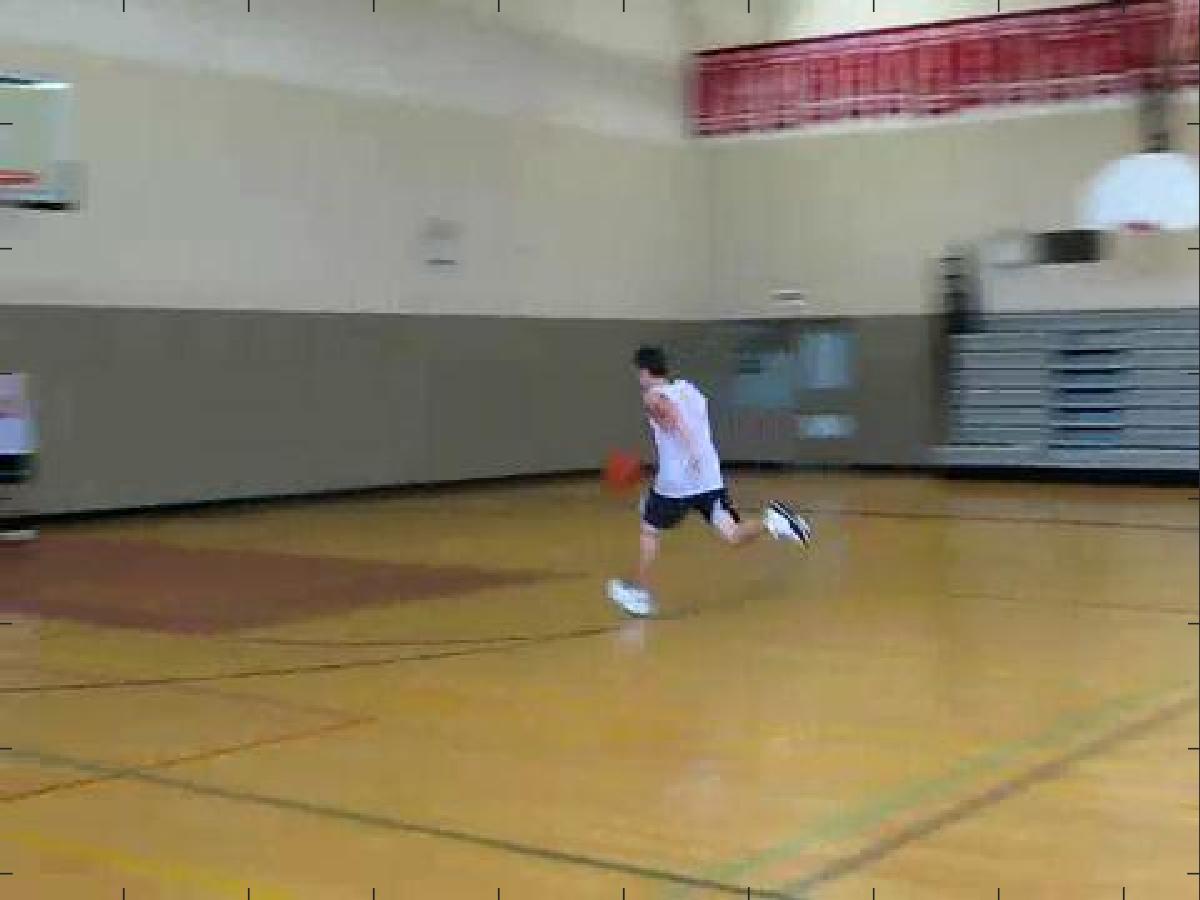}
c)\includegraphics[width=0.22\textwidth]{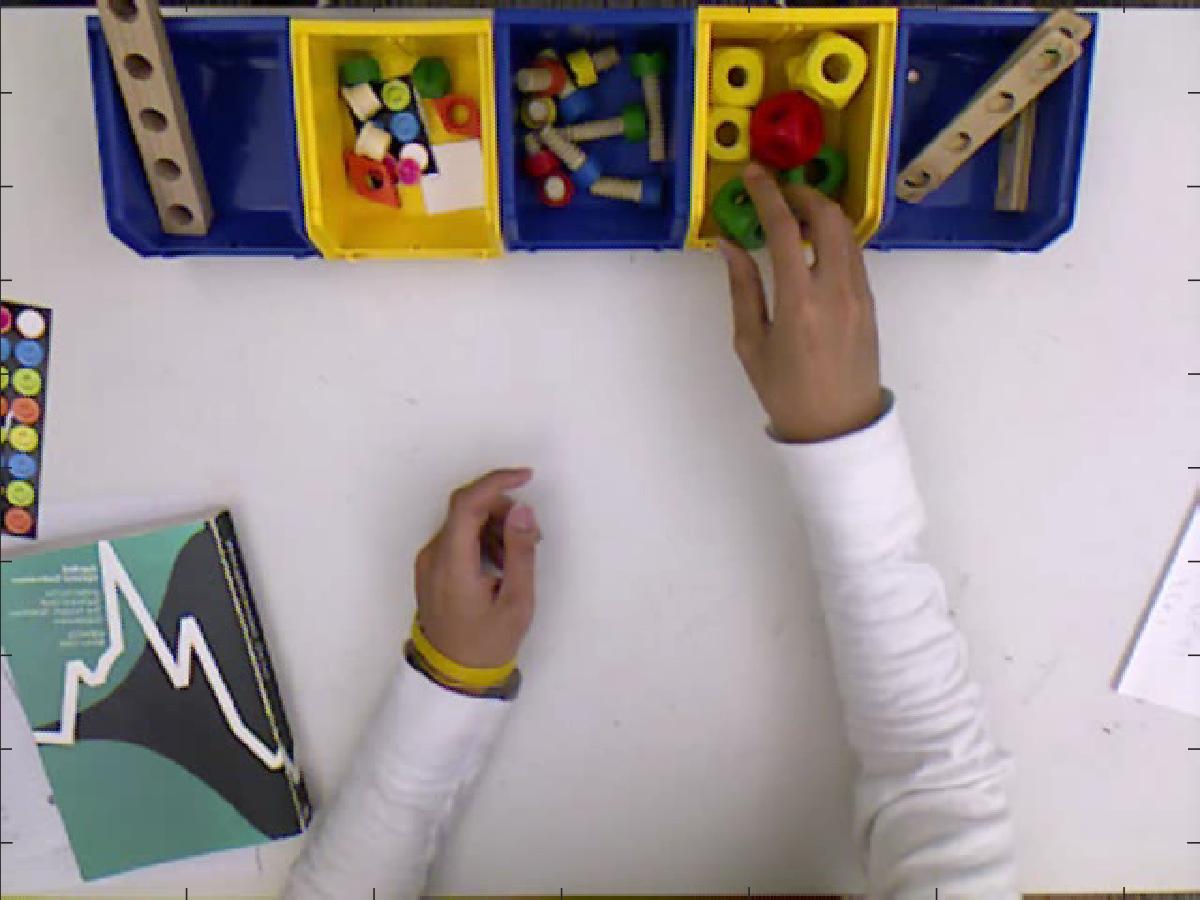}
d)\includegraphics[width=0.22\textwidth]{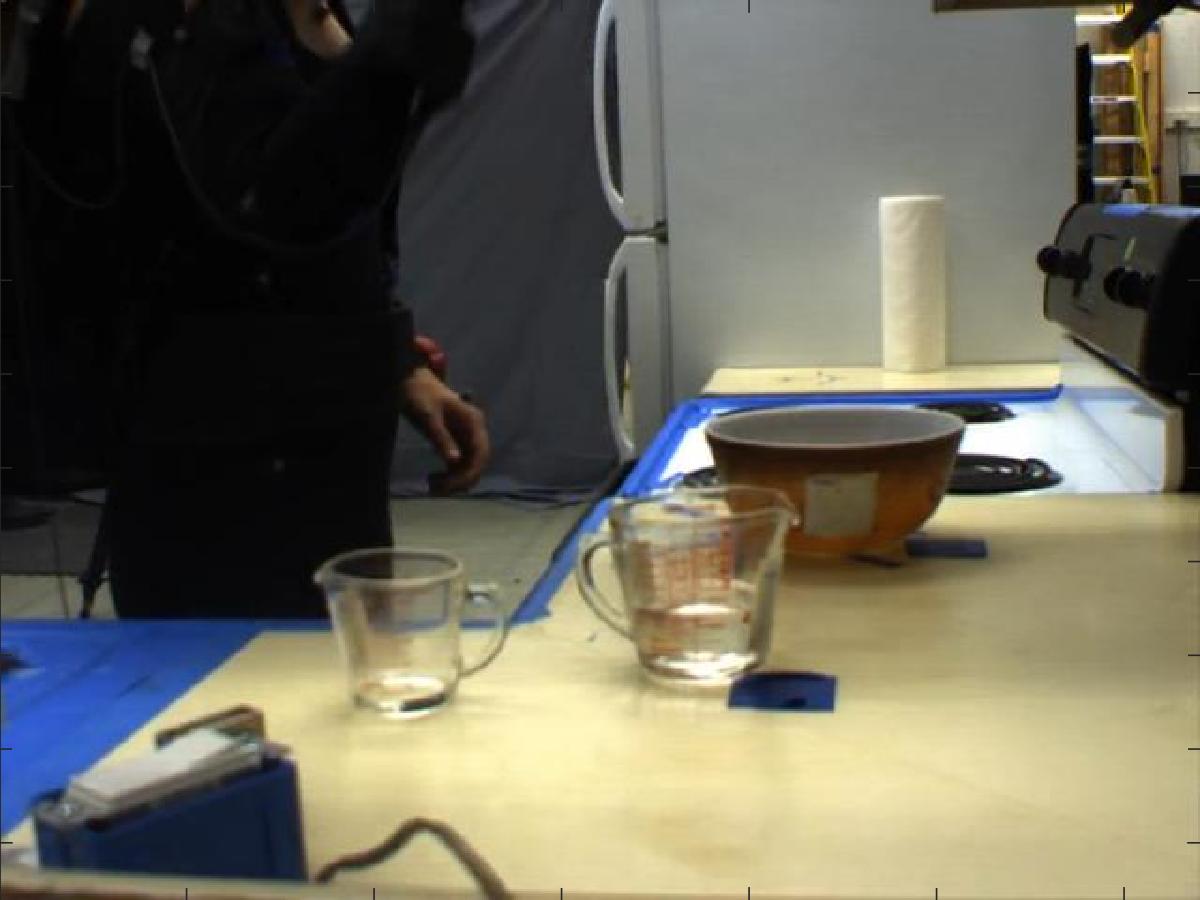}\\
e)\includegraphics[width=0.22\textwidth]{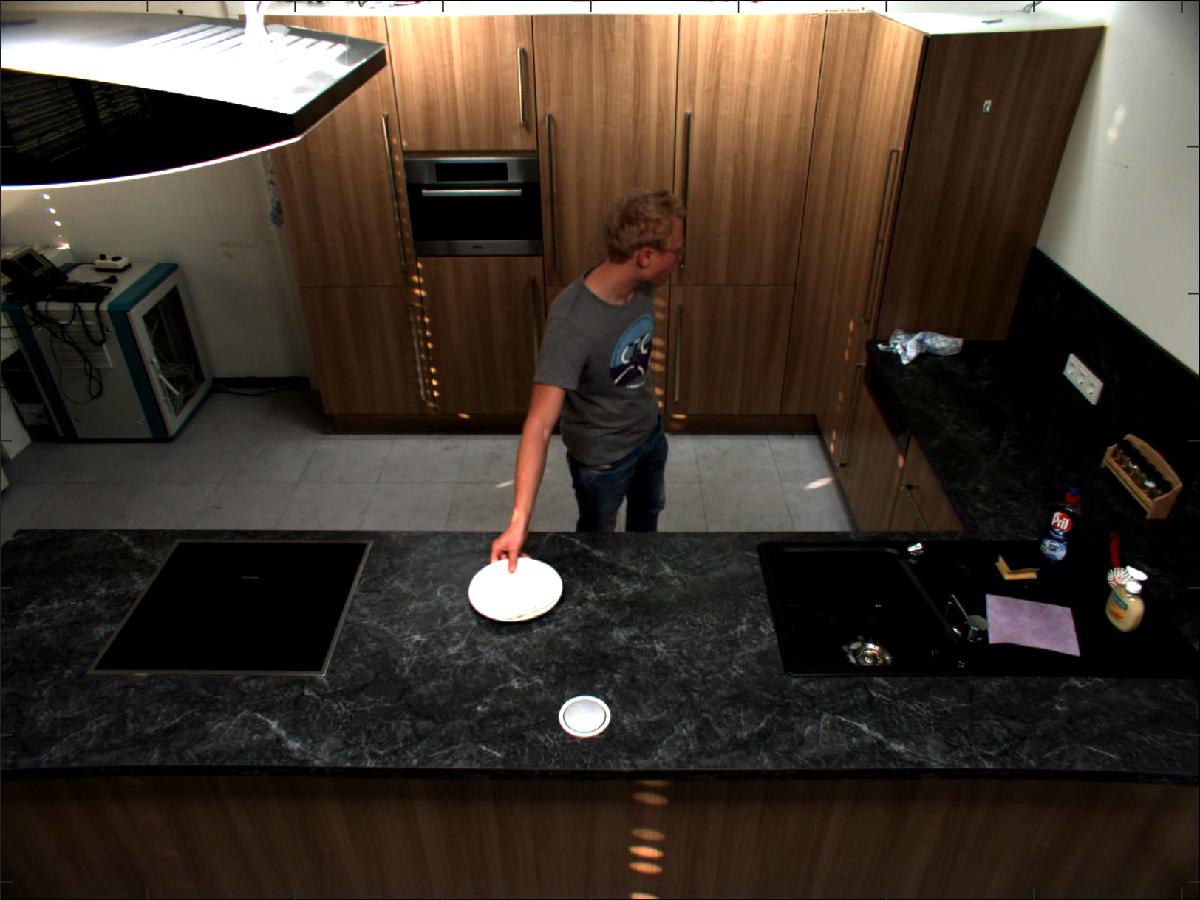}
f)\includegraphics[width=0.22\textwidth]{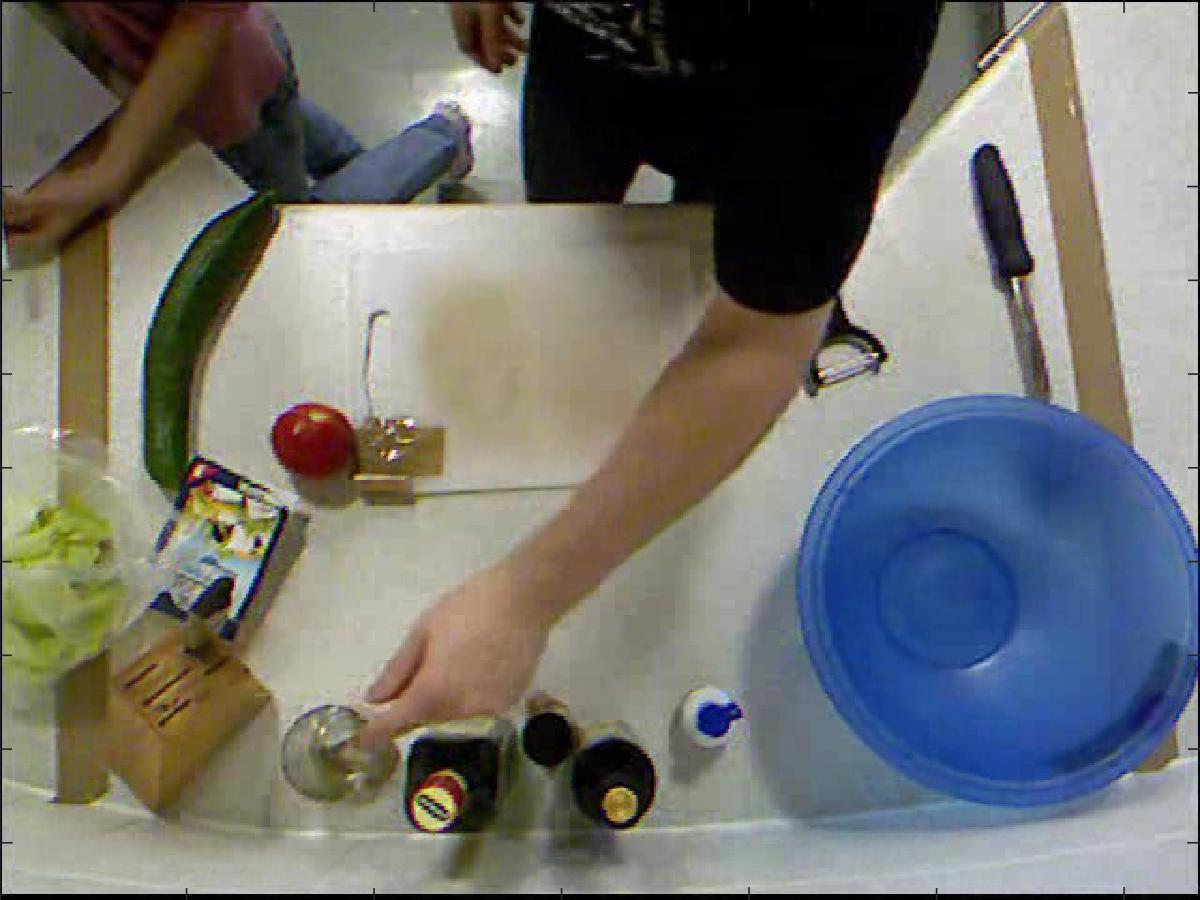}
g)\includegraphics[width=0.22\textwidth]{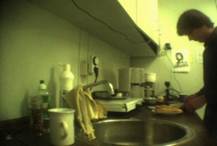}
h)\includegraphics[width=0.22\textwidth]{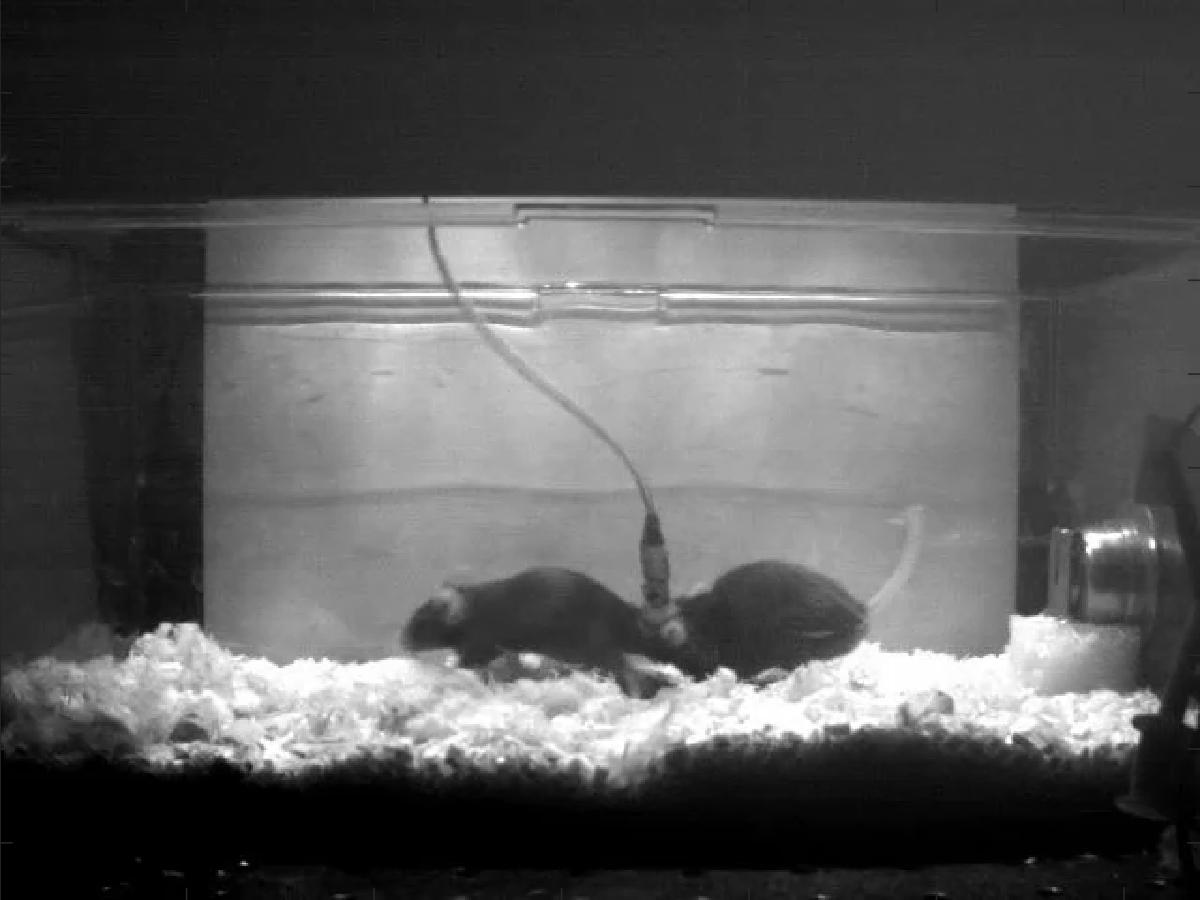}\\
\caption{Sample frames from the datasets used for performance evaluation: a) ADL~\cite{Messing2009Activity}, b) Olympic~\cite{Niebles2010Modeling}, c) ToyAssembly~\cite{Vo2014Stochastic}, d) CMU-MMAC~\cite{Spriggs2009TemporalSegmentation}, e) MPIICooking~\cite{Rohrbach2012database}, f) 50Salads~\cite{Stein2013Combining}, g) Breakfast~\cite{Kuehne2014Language}, and h) CRIM13~\cite{Burgos12CRIM}.}
\label{fig:dataset_samples}
\end{figure*}

Recent years have seen a significant increase in the availability of public activity datasets. To evaluate the proposed architecture, we considered complex activity datasets (as opposed to single task-oriented action) that are labeled at one or more levels of granularity. The datasets found suitable for this evaluation included: ADL~\cite{Messing2009Activity}, Olympics~\cite{Niebles2010Modeling}, ToyAssembly~\cite{Vo2014Stochastic}, CMU-MMAC~\cite{Spriggs2009TemporalSegmentation}, MPIICooking~\cite{Rohrbach2012database}, 50Salads~\cite{Stein2013Combining}, Breakfast~\cite{Kuehne2014Language}, and CRIM13~\cite{Burgos12CRIM}. Sample frames for each of these datasets are shown in Figure~\ref{fig:dataset_samples}. 

The recognition tasks for these datasets typically include activity classification, action unit detection and segmentation. The only exception is the Olympic Sport dataset, where no action unit labeling exists. For this dataset, we manually labeled 10 clips per class  and used these annotations for initializing the system. We then applied the recognition scheme to the remaining training clips and used the system outputs as labels for the training phase. 

Some of the selected datasets provide additional benefits such as multi-modal signals or multi-view settings. For this evaluation however, we only considered video data. All videos were separately processed and evaluated and we did not apply any method for combining camera input from different views. The duration of the datasets and the number of samples used for training is shown in Table~\ref{tab:overview_datasets}.

\begin{table}[t!]
   \centering
\begin{tabular}{|c|c|c|}
\hline 
               & Duration  & Train samples used per class \\ 
\hline 
 ADL           & 40 min 	  	 & 12-30 samples   \\
 Olympics      & 90 min 	     & 70-80 samples   \\
 Toy           & 64 min 	 	 & 15-20 samples \\ 
 CMU           & 265 min 	  	 & 30-40 samples   \\ 
 MPII          & 490 min 	   	 & 12-30 samples    \\ 
 50Salads      & 320 min 	 	 & 30-35 samples    \\
 BF            & 66.7 h 	  	 & 50-70 samples  \\
 CRIM13	       & 32.4 h 	     & 80-100 samples    \\
\hline 
\end{tabular} 
\caption{Overall duration of the different datasets and number of samples available for training.}
\label{tab:overview_datasets}
\end{table} 

\subsection{System evaluation}
\label{sec:System_evaluation}

We first compare the accuracy of the proposed reduced FVs against that of our previous work using HTK in combination with HOGHOF for the Breakfast dataset (Table~\ref{tab:recog_FV_SVM_HTK}). Replacing HOGHOF with DTFs already improves the overall system accuracy by $\sim10-14\%$ (Table~\ref{tab:recog_FV_SVM_HTK}, HTK+HOGHOF~w~PCA compared to HTK+DTF~w~PCA).

\begin{table}[t]\scriptsize
   \centering
\begin{tabular}{|c|c c c c c c|}
\hline 
\multicolumn{7}{ |c| }{Breakfast dataset - FV} \\ 
\hline 
      & GMMs = & 16 &  32 &  64 & 128 &  256 \\ 
\hline 
1) SVM+DTF w/o PCA &      & 52.0  &  52.6  &  48.7  &  39.6  &  23.2    \\ 
2) SVM+DTF  w PCA & $D'=64$ & 42.0  &  42.5  &  42.8  &  40.3  &  41.2   \\ 
\hline 
3) HTK+HOGHOF w PCA & $D'=64$ &  62.3 &   61.1 &   62.2  &  60.7  &  60.2  \\ 
4) HTK+DTF w PCA   & $D'=64$ &  \textbf{71.5} &  \textbf{72.2}  & \textbf{73.3} &  \textbf{68.6}   & \textbf{66.4}    \\  

\hline 
\end{tabular} 
\caption{Comparison between HTK vs. SVM and HOGHOF vs. DTFs for activity recognition (in combination with FV-based encoding on the Breakfast dataset).}
\label{tab:recog_FV_SVM_HTK}
\end{table}

\begin{table*}[t] 
   \centering
\begin{tabular}{|c|l|l|l|l|l|l|l|l|}
\hline 
\multicolumn{9}{ |c| }{Segmentation } \\ 
\hline 
GMM= & ADL          & Oly.          & Toy                & CMU               & MPII               & 50Salad      & BF                         & CRIM13 \\ 
\hline 
16    & 53.4         & 62.4          & 50.3 / \textit{64.3}    & 53.8 / \textit{60.8}     & 46.5 / \textit{58.5}          & 81.6           & 36.2 / \textit{54.2}      & 52.6   \\ 
32    & 54.5         & 66.1          & 48.6 / \textit{63.1}    & 53.7 / \textit{60.7}     & 53.9 / \textit{68.5}          & 80.4             & 36.9 / \textit{54.4} &\textbf{53.5} \\ 
64    & 55.7         & 67.5          & 56.7 / \textit{67.5}    & 53.0 / \textit{60.3}     & 51.6 / \textit{63.9}          & \textbf{83.8}  & \textbf{38.1 / \textit{56.3}}& 53.4   \\ 
128   & 58.9         & 65.9          & 60.5 / \textit{70.8}    & 52.5 / \textit{60.4}     & 53.9 / \textit{66.8}          & 82.0           & 34.0 / \textit{51.2}         & 52.6   \\ 
256   & 54.4         & 63.7          & 63.5 / \textit{72.2}    & \textbf{58.8 / \textit{67.1}} & \textbf{57.3 / \textit{71.7}} & \textbf{83.8}  & 32.7 / \textit{50.7}     &  53.3   \\ 
\hline 
Best    & \hspace{0.2cm}--\hspace{0.2cm}  & \hspace{0.2cm}--\hspace{0.2cm} & \hspace{0.2cm}--\hspace{0.2cm} / \textbf{\textit{91.0}}~\cite{Vo2014Stochastic} & \hspace{0.2cm}--\hspace{0.2cm} / \textit{59.0}~\cite{Vo2014Stochastic} &  \hspace{0.2cm}--\hspace{0.2cm} / \textit{54.3}~\cite{Ni2014Multiple}    & 67.6~\cite{Stein2013Combining}     & \hspace{0.2cm}--\hspace{0.2cm} / \textit{28.8}~\cite{Kuehne2014Language}     & 39.1~\cite{Burgos12CRIM} \\ 
\hline 
\end{tabular} 
\caption{Overview of the segmentation results for all datasets. Accuracy is computed as the mean over all classes. For comparison, we also report the frame-based accuracy (\textit{italic}) for the Toy, CMU and BF dataset, and midpoint hit accuracy (\textit{also italic}) for the MPII dataset as used by the authors in the original studies.}
\label{tab:eval_dataset_segmentation}
\end{table*}

To evaluate the impact of the reduced FVs on a generative vs. a discriminative framework, we compared the proposed pipeline against one where the HTK classification stage was replaced with an SVM (for both the full FV representation with 2,048--32,768 dimensions and K= 16--256 GMMs and the reduced FV representation with 64 dimensions). Classification was based on the libSVM software library~\cite{Chang2011LIBSVM} using a linear kernel. We used identical features and GMM clusters as in the proposed HTK-based system. Note, however, that for the SVM baseline, features were sampled from the entire video sequence because we found it to work better than a frame-based sampling as used for HTK. 

As Table~\ref{tab:recog_FV_SVM_HTK} shows, SVM-based classification performs better when using the full FV representation for classification compared to the reduced FV representation. However, the accuracy of the SVM-based classification remains significantly below the accuracy of the system based on HTK by $\sim20-30\%$ with identical features. Our results show that, compared to the baseline reported in~\cite{Kuehne2014Language}, reduced FVs improve the recognition accuracy by $\sim20\%$ for HOGHOF and $\sim30\%$ for DT. 

\subsection{Segmentation}
\label{sec:Segmentation}

To evaluate the segmentation accuracy of the proposed system, we consider eight different datasets (see section~\ref{sec:Datasets} for details). As the original benchmarks for these datasets are based on different measurements, we report multiple accuracy measures for fair comparison to these baseline systems. One measure reported uses the mean accuracy over all classes (corresponding to the mean accuracy computed over the diagonal of the corresponding confidence matrix) as used in~\cite{Rohrbach2012database, Stein2013Combining, Burgos12CRIM}. In addition, we also report the frame-based accuracy (corresponding to the mean proportion of correctly classified frames) for the Toy, CMU and Breakfast dataset as used in~\cite{Vo2014Stochastic,Kuehne2014Language}. For the MPII dataset, we also report the mid-point hit accuracy as defined in~\cite{Rohrbach2012database}. 

\begin{figure*}[t!]
\centering
a)\includegraphics[width=0.3\linewidth]{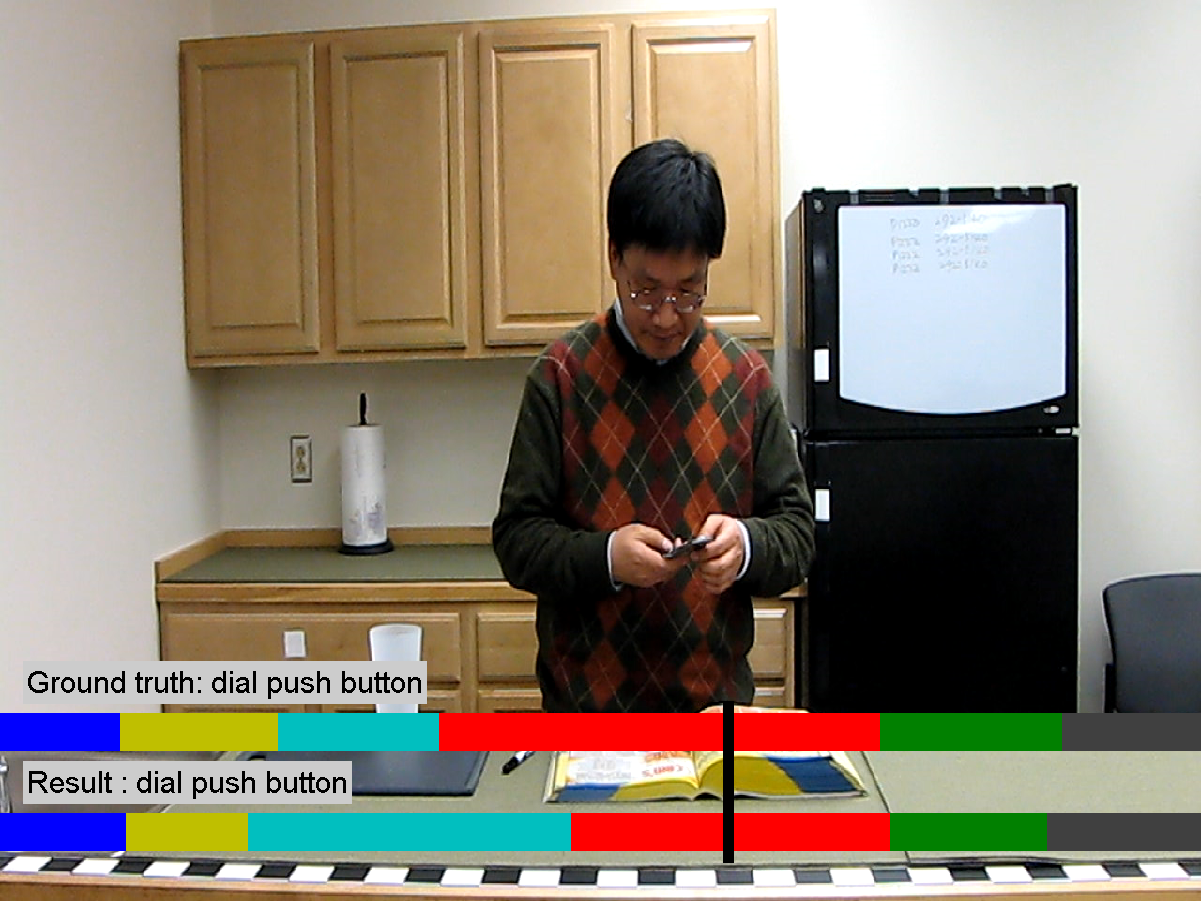}   \hspace{1mm}
b)\includegraphics[width=0.3\linewidth]{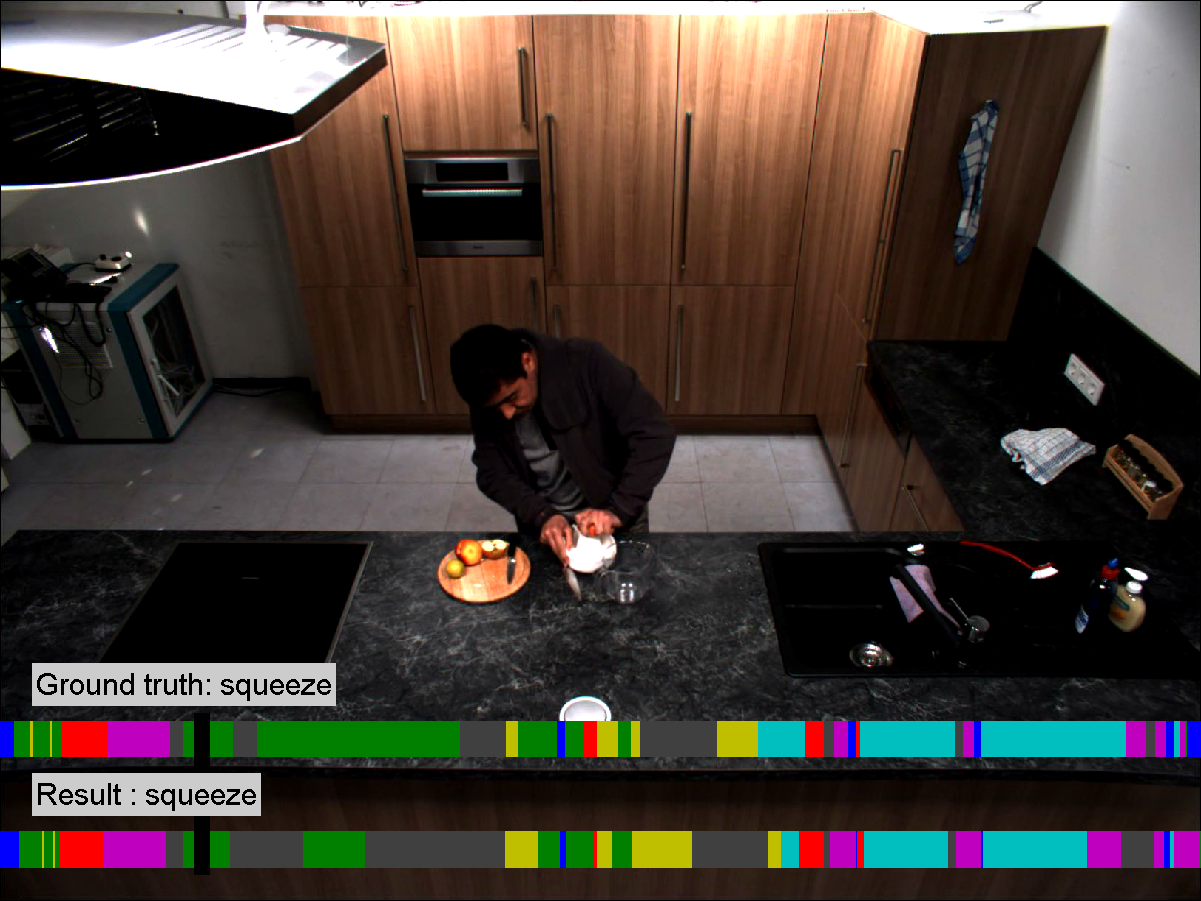} \hspace{1mm}
c)\includegraphics[width=0.3\linewidth]{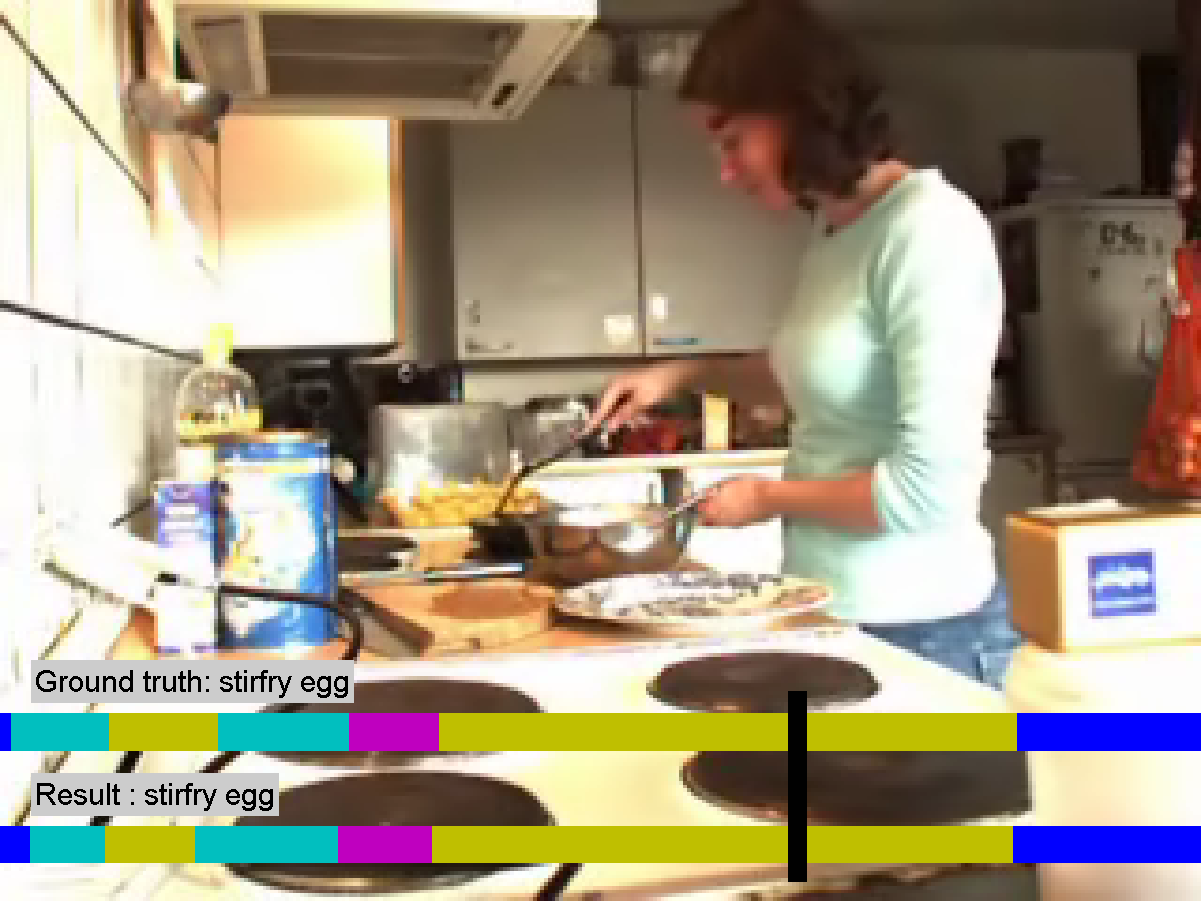}   \\
\caption{Sample segmentation results for a) the ADL dataset (``dial phone''), b) the MPII cooking dataset (``prepare cold drink''), and c) the Breakfast dataset (``prepare scrambled eggs''). The upper/lower color bars correspond to ground-truth/system outputs, respectively.}
\label{fig:exp_frame_recog}
\end{figure*}

Segmentation results for the proposed system and available benchmarks are reported in Table~\ref{tab:eval_dataset_segmentation}. Note that for the two smallest datasets (ADL and Olympics), no benchmark is available as no segmentation results have been previously reported for these datasets. It is pretty clear that the proposed approach under-performs the best segmentation results obtained for the Toy assembly dataset (which remains a small video dataset with about one hour of video). 
For large datasets (8 hours or more of video), the system significantly outperforms the state of the art in terms of segmentation accuracy (\eg  BF +27.5\%). Note that for the CRIM13 dataset, the benchmark approach is based on spatial-temporal features~\cite{Burgos12CRIM}. For this evaluation, we only considered side-view videos and report the accuracy of the benchmark system for the same set of videos as reported in the original study. Sample segmentation results are shown in Figure~\ref{fig:exp_frame_recog} for the ADL, MPII and Breakfast datasets. 

\enlargethispage{.3cm}
\subsection{Activity classification}
\label{sec:Activity}

Here, we evaluated the accuracy of the proposed system for activity classification (Table~\ref{tab:eval_datasets}). We only considered datasets that provide multiple activity classes (\ie ADL, Olympic and Breakfast datasets). Consistent with earlier experiments, the accuracy of the proposed system is below the state of the art for smaller datasets (\eg ADL and Olympic Sports) but outperforms the state of the art when enough training samples are available (\eg Breakfast dataset).  

\begin{table}[t!]
   \centering
\begin{tabular}{|c|c|c|c|}
\hline 
\multicolumn{4}{ |c| }{Activity classification } \\ 
\hline 
GMM=  & ADL & Olympics &  BF \\ 
\hline 
16    & 86.0  & 74.4            &71.5  \\ 
32    & 86.7  & 76.8            &72.2  \\ 
64    & 91.3  & 77.6            &\textbf{73.3}  \\ 
128   & 94.7  & 77.2             &68.6  \\ 
256   & 87.3  & 74.4              &66.4  \\ 
\hline 
\hline 
Best & \textbf{98.7}~\cite{Rostamzadeh2013Daily} & \textbf{90.2}~\cite{Wang2013Action} &  40.5~\cite{Kuehne2014Language} \\ 
\hline 
\end{tabular} 
\caption{Activity classification results.}
\label{tab:eval_datasets}
\end{table}

\section{Conclusion}

In this paper, we studied how different feature representations affect the performance of a structured generative (temporal) model based on the HTK framework. We performed a systematic evaluation of the proposed approach and compared the accuracy of the resulting system against the state of the art for both activity segmentation and classification. Our results showed that combining a compact video representation based on Fisher Vectors with Hidden Markov Models yields very significant gains in accuracy for both the recognition of goal-oriented activities and their parsing at the level of task-oriented action units. Indeed, when sufficient training data was available, we found that structured generative temporal models outperform the state of the art. These results are consistent with recent trends in other areas of computer vision suggesting that, as datasets are becoming increasingly large, structured models are starting to outperform the state of the art. 

\section{Acknowledgment}
This work was supported by the DFG Emmy Noether program (GA 1927/1-1) to HK and by DARPA young faculty award (N66001-14-1-4037) and NSF early career award (IIS-1252951) to TS.

\balance
{\small
\bibliographystyle{ieee}
\bibliography{bibliography}
}

\end{document}